\documentclass{article}

\usepackage{arxiv}

\usepackage[utf8]{inputenc} 
\usepackage[T1]{fontenc}    
\usepackage{hyperref}       
\usepackage{url}            
\usepackage{booktabs}       
\usepackage{amsfonts}       
\usepackage{nicefrac}       
\usepackage{microtype}      
\usepackage{lipsum}
\usepackage{graphicx}
\usepackage{amsmath}
\usepackage{multirow, array}

\usepackage[frozencache, cachedir=minted-cache]{minted}

\graphicspath{ {./figures/} }
\usepackage[skip=2pt,font=small]{caption}

\usepackage{authblk}

\usepackage[numbers]{natbib}
\usepackage{lipsum}
\usepackage{subcaption}
\usepackage{enumitem}

\usepackage{framed} 
\usepackage{multicol} 
\usepackage{nomencl} 
\makenomenclature

\providecommand{\keywords}[1]
{
  \small	
  \textbf{\textit{Keywords---}} #1
}

\title{DeepTSF: Codeless machine learning operations for time series forecasting}

\author{Sotiris Pelekis}
\author{Evangelos Karakolis}
\author{Theodosios Pountridis}
\author{George Kormpakis}
\author{George Lampropoulos}
\author{Spiros Mouzakits}
\author{Dimitris Askounis}

\affil{Decision Support Systems Laboratory \\
  School of Electrical and Computer Engineering\\
  Institute of Communications and Computer Systems\\ National Technical University of Athens\\
  Greece\\}

\begin{document}
\maketitle
\begin{abstract}
This paper presents DeepTSF, a comprehensive machine learning operations (MLOps) framework aiming to innovate time series forecasting through workflow automation and codeless modeling. DeepTSF automates key aspects of the ML lifecycle, making it an ideal tool for data scientists and MLops engineers engaged in machine learning (ML) and deep learning (DL)-based forecasting. DeepTSF empowers users with a robust and user-friendly solution, while it is designed to  seamlessly integrate with existing data analysis workflows, providing enhanced productivity and compatibility. The framework offers a front-end user interface (UI) suitable for data scientists, as well as other higher-level stakeholders, enabling comprehensive understanding through insightful visualizations and evaluation metrics. DeepTSF also prioritizes security through identity management and access authorization mechanisms. The application of DeepTSF in real-life use cases of the I-NERGY project has already proven DeepTSF's efficacy in DL-based load forecasting, showcasing its significant added value in the electrical power and energy systems domain.
\end{abstract}

\keywords{automation, codeless, deep learning, machine learning operations, time series forecasting, software}



\begin{table*}[!t]   
\begin{framed}
\nomenclature{\textbf{ANN}}{Artificial neural network}
\nomenclature{\textbf{API}}{Application programming interface}
\nomenclature{\textbf{CLI}}{Command line interface}
\nomenclature{\textbf{CNN}}{Convolutional neural network}
\nomenclature{\textbf{DL}}{Deep learning}
\nomenclature{\textbf{GPU}}{Graphics processing unit}
\nomenclature{\textbf{LSTM}}{Long short-term memory}
\nomenclature{\textbf{LW}}{Lookback window}
\nomenclature{\textbf{MAPE}}{Mean absolute percentage error}
\nomenclature{\textbf{MLOps}}{Machine learning operations}
\nomenclature{\textbf{MLP}}{Multi-layer perceptron}
\nomenclature{\textbf{MLR}}{Multiple linear regression}
\nomenclature{\textbf{N-BEATS}}{Neural basis expansion analysis for time series forecasting}
\nomenclature{\textbf{RNN}}{Recurrent neural network}
\nomenclature{\textbf{RMSE}}{Root mean squared error}
\nomenclature{\textbf{STLF}}{Short-term load forecasting}
\nomenclature{\textbf{sNaive}}{Seasonal naive}
\nomenclature{\textbf{TCN}}{Temporal convolutional network}
\nomenclature{\textbf{UI}}{User interface}
\printnomenclature
\end{framed}
\end{table*}

\section{Motivation and significance} \label{sec:1}


Historically, time-series modeling has been a prominent area of interest in academic research, with diverse applications in fields such as climate modeling \citep{Mudelsee2019TrendMethods}, biological sciences \citep{Stoffer2012Editorial:Sciences}, medicine \citep{Topol2019High-performanceIntelligence}, and commercial decision-making domains like retail \citep{Spiliotis2021ProductData}, finance \citep{Sezer2020Financial20052019}, and energy \citep{Pelekis2023ADrivers, Pelekis2022InPerformance}. Traditional approaches in this field have primarily focused on parametric statistical models, utilizing domain expertise-driven techniques such as autoregressive models \citep{BoxG.1976TimeControl}, exponential smoothing \citep{Gardner1985ExponentialArt}, and other methods that heavily relied on decomposing time series \citep{Assimakopoulos2000TheForecasting}. However, the advent of modern ML methods has introduced data-driven approaches for capturing temporal dynamics \citep{Masini2023MachineForecasting}. 

Among these methods, deep learning (DL) has gained significant traction, inspired by its remarkable achievements in areas like image classification \citep{Krizhevsky2017ImageNetNetworks}, natural language processing \citep{Young2018RecentProcessing}, and reinforcement learning \citep{Li2017DeepOverview}. Deep neural networks, with their customized architectural assumptions or inductive biases \citep{Baxter2000ALearning}, can effectively learn intricate data representations, eliminating the need for manual feature engineering and model design. The availability of open-source backpropagation frameworks \citep{Paszke2019PyTorch:Library, Abadi2016TensorFlow:Scale} has further simplified network training, allowing for flexible customization of network components and loss functions. This convergence of data availability, increased computing power, and the rise of deep learning has transformed time-series forecasting, paving the way for the next generation of models. These models leverage ML techniques to effectively capture complex temporal dependencies, facilitating improved forecasting accuracy and flexibility in various applications.

From an implementation and development perspective, today's software systems are wide open to innovative approaches enabling the incorporation of real-time and data-centric solutions powered by information and communication technologies (ICT) such as big data, internet of things (IoT), artificial intelligence (AI), and lately machine learning operations (MLOps) \citep{Alla2021}. MLOps is an emerging field in AI that establishes a new paradigm in industrial ML that goes beyond conventional, and manual model development processes. In this context, automated and continuous model training, evaluation, validation, and deployment already replace the manual processes of the conventional ML lifecycle which is the core of current time series forecasting pipelines, leading to faster and more efficient decision making for related stakeholders.

Specifically, Python provides a diverse selection of open-source time series forecasting tools including PyTorch-Forecasting \citep{PyTorch2023PyTorchForecasting}, GluonTS \citep{Alexandrov2019GluonTS:Python}, Prophet \citep{Prophet2023ProphetScale.}, Darts \citep{Herzen2021Darts:Series}, and other relevant options. PyTorch-Forecasting is a Python library built on the PyTorch framework, specifically designed for deep learning-based time series forecasting. It provides a high-level API and supports advanced neural network architectures, enabling users to leverage the power of deep learning for accurate predictions. PyTorch-Forecasting is well-regarded for its flexibility and customization options, making it a preferred choice for researchers and practitioners with expertise in deep learning. GluonTS, built on the Apache MXNet framework \citep{Apache2023ApacheLearning.}, is a library focused on probabilistic time series forecasting. It offers a comprehensive suite of tools and models for capturing uncertainty and modeling complex patterns in time series data. GluonTS provides state-of-the-art probabilistic forecasting models such as DeepAR, RNN-based models, and Temporal Convolutional Networks (TCN). Additionally, it offers utilities for data preprocessing, model evaluation, and visualization.
Prophet, developed by Facebook, is a widely-used library specifically designed for time series forecasting. It offers a user-friendly interface and employs an additive model that decomposes time series into trend, seasonality, and holiday components. Prophet simplifies the forecasting process by automating trend detection and supporting custom regressors. Its simplicity and ease of use make it accessible to users with varying levels of expertise in time series forecasting. Darts is a versatile library for time series forecasting. It provides a comprehensive collection of models, including classical statistical models, machine learning algorithms, and deep learning architectures. Darts emphasizes automation and MLOps capabilities, offering automated hyperparameter optimization and model selection. It also seamlessly integrates with other Python libraries, facilitating efficient data preprocessing and feature engineering. In addition to these tools, there are other notable time series forecasting libraries available in Python. For example, scikit-learn \citep{Pedregosa2011Scikit-learn:Python}, a popular machine learning library, offers various algorithms suitable for time series forecasting along with utilities for model evaluation and selection. TensorFlow Probability \citep{TensorFlow2023TensorFlowProbability} combines deep learning with probabilistic modeling, providing a powerful framework for capturing uncertainty in predictions.

In this paper we present DeepTSF, a full-stack machine learning operations (MLOps) framework that provides codeless machine learning (ML) capabilities for time series forecasting by automating several parts of the ML lifecycle. The back-end of DeepTSF
is specifically developed to automate the ML and DL-based forecasting lifecycle for data scientists and MLops engineers. It incorporates state-of-the-art ML and DL algorithms and techniques, providing them with a user-friendly and powerful tool for accurate and efficient time series forecasting. DeepTSF harnesses the capabilities of Python's extensive scientific libraries and frameworks, enabling seamless integration with existing data analysis workflows. Additionally, it provides a front-end that can serve several other categories of high-level stakeholders and end-users within the time series modeling domain by providing high-level insights--through visualizations and evaluation metrics-- of the forecasting models developed at lower level by the engineers. Ultimately, securing the provided services is a high priority of DeepTSF and is achieved by employing effective identity management and access authorization mechanisms. In terms of real-world application, DeepTSF has already been validated by researchers \citep{Pelekis2023ADrivers, Pelekis2022InPerformance} within DL-based short-term electricity load forecasting use cases. However, in practice, DeepTSF can handle all types of forecasting tasks irrespective of the time series domain, resolution, and forecast horizon.

\section{Software description}
\label{sec:2}

DeepTSF is a user friendly, AI-based time series forecasting application that facilitates MLOps for multi-horizon time series forecasting with deep learning models. DeepTSF has been developed using various programming tools in Python and Javascript alongside Docker \cite{Docker2023Docker:Development} for efficient building, sharing and deployment of the application.

\subsection{Software Architecture} \label{sec:2:architecture}

\begin{figure}[htb!]
	\centering
        \includegraphics[width=\textwidth, keepaspectratio]{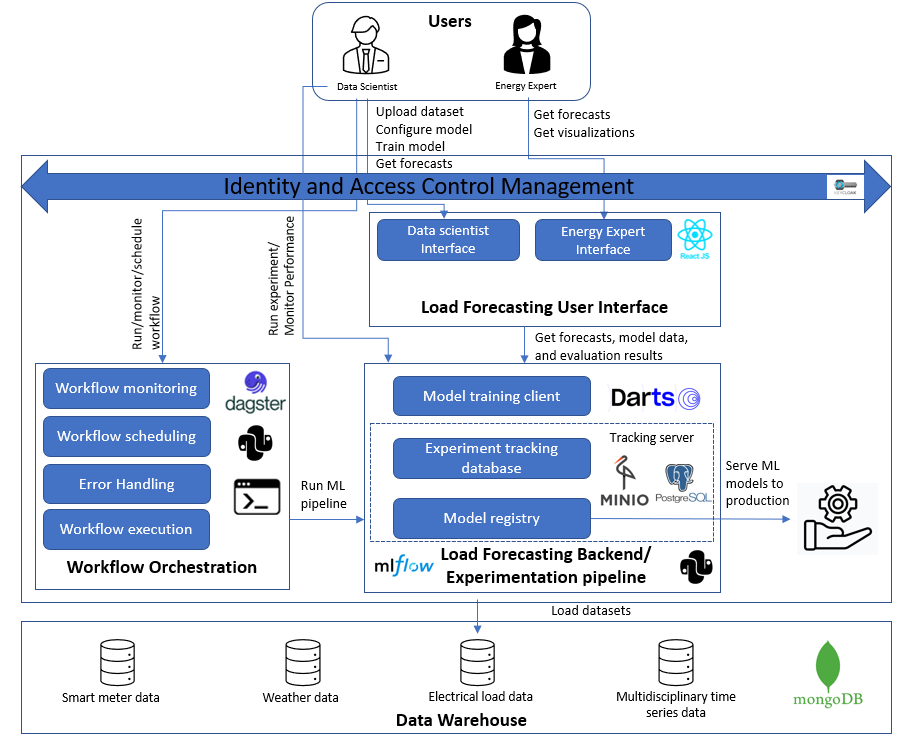}
	  \caption{The DeepTSF architecture}\label{fig:architecture}
\end{figure}

For the development of DeepTSF, a modular and micro-service-oriented approach has been followed, consisting of several layers that assume different responsibilities and implement various functionalities within the application. In this context, the overall DeepTSF architecture (also illustrated in Fig. \ref{fig:architecture}) consists of several layers that employ state-of-the-art programming frameworks as follows.

\begin{itemize}
\item \textbf{Database}: Time series data (e.g. energy and weather-related series data) are periodically stored to a MongoDB \citep{MongoDB2023MongoDB:MongoDB} (NoSQL database) after several pre-processing and data harmonization processes. MongoDB was selected because it provides great performance and scalability, as it is able of processing millions of requests per day without changes in performance. Furthermore, it provides flexibility to the developers, in the sense that stored objects do not need to follow a strict data model and they map directly to objects in the code. Additionally, it provides the ability to embed related data to a document, to increase performance and reduce computational costs. Of course, there is a wide variety of (big) data storage technologies that serve different requirements. To this end, we aim to abstract the data storage layer, as DeepTSF aims to support more data storage technologies in the future by giving the users the ability to connect their own databases.

\item \textbf{Forecasting backend:} This component is the core of DeepTSF and is responsible for: i) extracting data from the database, ii) preprocessing and harmonization in accordance with the specifications of the forecasting model input, iii) model training and (optionally) hyperparameter tuning, iv) model evaluation and v) model storage, versioning and serving. These steps can be executed either sequentially or individually. The execution/orchestration of the pipeline can take place either through CLI or the workflow orchestrator that is described later on. 
\par
The backend's code is organized in files that correspond to the stages of the above workflow:
\begin{itemize}
    \item \textbf{load\_raw\_data.py}: This interface is responsible for loading the data from local or online sources and saving it to the MLflow tracking server.
    \item \textbf{etl.py}: Conducts data harmonization and pre-processing.
    \item \textbf{training.py}: Performs model training.
    \item \textbf{optuna\_search.py}: Responsible for the hyperparameter tuning of the model if requested by the user.
    \item \textbf{evaluate\_forecasts.py}: It evaluates the model using the provided test set.
\end{itemize}
The DeepTSF documentation \citep{DeepTSF2023DeepTSFDocumentation} includes more information on the backend's code structure and usage.

The most important technologies of the forecasting backend are the following: 
\begin{itemize}
    \item \textbf{Python MLflow} \citep{Alla2021} as the core of this implementation. The model training client's code extensively uses the MLflow API to allow for machine learning experiment triggering and tracking on the tracking server. Optionally, models can also be registered, versioned, and deployed using the MLflow model registry \citep{MLflow2023MLflowRegistry}. The tracking server is also assisted by a MinIO \citep{MinIO2022MinIOStorage} artifact store for storing models, datasets, and results alongside a PostgreSQL \citep{PostgreSQL2022PostgreSQLDatabase} database for storing experiment parameters and evaluation metrics. Note here that a separate docker deployment \citep{LampropoulosGeorge2023MLflowImplementation}, that was developed in parallel with DeepTSF, is required for the optional set up of the MLflow tracking server.
    \item \textbf{Python Darts} \citep{Paszke2019PyTorch:Library} for automated and optimized time series forecasting based on PyTorch.
    \item \textbf{Python Optuna} library \citep{Akiba2019Optuna:Framework} for hyperparameter optimization. 
    \item \textbf{Python SHAP (SHapley Additive exPlanations)} library \citep{Shap2023SHAPDocumentation} for explaining feature importance.
    \item \textbf{Python FastAPI} \citep{Lathkar2023GettingFastAPI} for exposing the information and functionalities of the back-end application as an API to the outside world.
\end{itemize}
Further details on the architectural elements of this subcomponent are provided in \ref{app:b}.

\item \textbf{Graphical user interface} which is the front-end environment of the DeepTSF is developed using React \citep{ArtemijFedosejev2015React.jsEssentials}, an open-source, widely-used and very well-documented JavaScript library, which offers versatility and unparalleled efficiency for building user interfaces. Its main advantages lie in the component-based architecture, which enables the developers to create reusable and modular UI components, resulting in faster rendering and optimal overall application performance. Moreover, React offers a robust ecosystem of libraries and tools, ensuring that the created applications will be scalable and maintainable, while there is also a very active community of developers that offers support and contributions to the React library.

\item \textbf{Workflow Management} DeepTSF supports two layers of workflow management, as follows.
\begin{itemize}
\item \textbf{Basic workflow management} The first layer comprises a CLI which has been developed with the Click package \citep{Click2023ClickDocumentation} and allows for executing (individually or sequentially) the four stages of the ML pipeline that were above mentioned. This CLI is complemented by the built-in CLI of MLflow \citep{MLflow2023Command-LineDocumentation}.

\item \textbf{Advanced workflow management} The advanced workflow management engine is based on Dagster \citep{Dagster2023DagsterPipelines}. This engine allows for orchestrating, scheduling and executing the ML pipeline in a more advanced fashion. This component is integrated with the MongoDB to extract the data needed for the forecasting scenario, as well as with the MLflow platform for tracking the experiments and storing the trained models. This integration is achieved utilizing the python client APIs of those tools. In addition, the GraphQL API of Dagster \citep{GraphQL2023GraphQLAPI} is utilized to expose functionality of executing the defined jobs to the end-users of application. Note here that every forecasting use case requires a new custom configuration of the Dagster orchestrator, which is abstracted within our work so that it can be used as a starting point for future extensions.
\end{itemize}

\item \textbf{Identity and access management}
Identity and access management in DeepTSF are implemented leveraging OpenID Connect Protocol and therefore the OAuth2.0 protocol on top of which OpenID Connect is built \citep{Fett2017TheGuidelines}. A Keycloak \citep{Camposo2021SecuringKeycloak} server is used and configured to act as the OpenId Provider. User authentication is achieved via the front-end application, where users authenticate against the OpenID Provider using their credentials. After successful authentication, a token exchange  request takes place, resulting to access token issuance and its exchange. Concerning the back-end service of the application, the access to its FastAPI endpoints is validated against OpenID Provider by utilizing its introspection endpoint providing the access token sent within user's request and by role validation logic implemented in FastAPI. Both front-end and back-end services are registered within the OpenID Provider as confidential clients in order to establish a trusted relationship. A reverse proxy setup has been developed to enable authentication for the MLflow server. This reverse proxy service is an OpenResty \citep{OpenResty2023OpenRestyLuaJIT} webservice which extends the NGINX \citep{Reese2008Nginx:Proxy} with Lua and act as the OpenID Connect Relying Party authenticating the users against the OpenID Provider.   
\end{itemize}

\subsection{Software Functionalities} \label{sec:2:functions}

\subsubsection{User interface} \label{sec:2:ui}
The front-end of DeepTSF is designed to provide a smooth experience to the end-user roles within the application (system administrators, data scientists and domain experts). The following sub-sections describe the main functionalities of the DeepTSF front-end environment. More precisely, each page of the front-end application will be presented, along with a detailed description of the inputs and the outputs, as well as the overall functionalities it offers.

\par
\begin{figure*}[htb!]
\centering
\begin{subfigure}[b]{.35\linewidth}
   \centering
   \includegraphics[height=4cm, width=\textwidth]{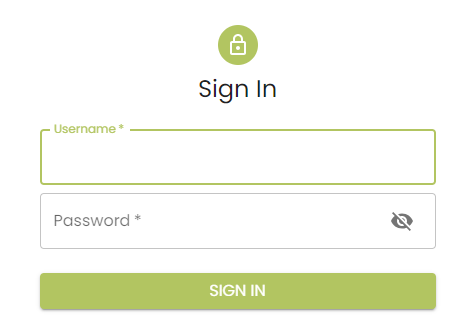}
   \caption{DeepTSF Sign-In page.}
   \label{fig:signin}
\end{subfigure}
\hfill
\begin{subfigure}[b]{.6\linewidth}
   \centering
   \includegraphics[height=4cm, width=\textwidth]{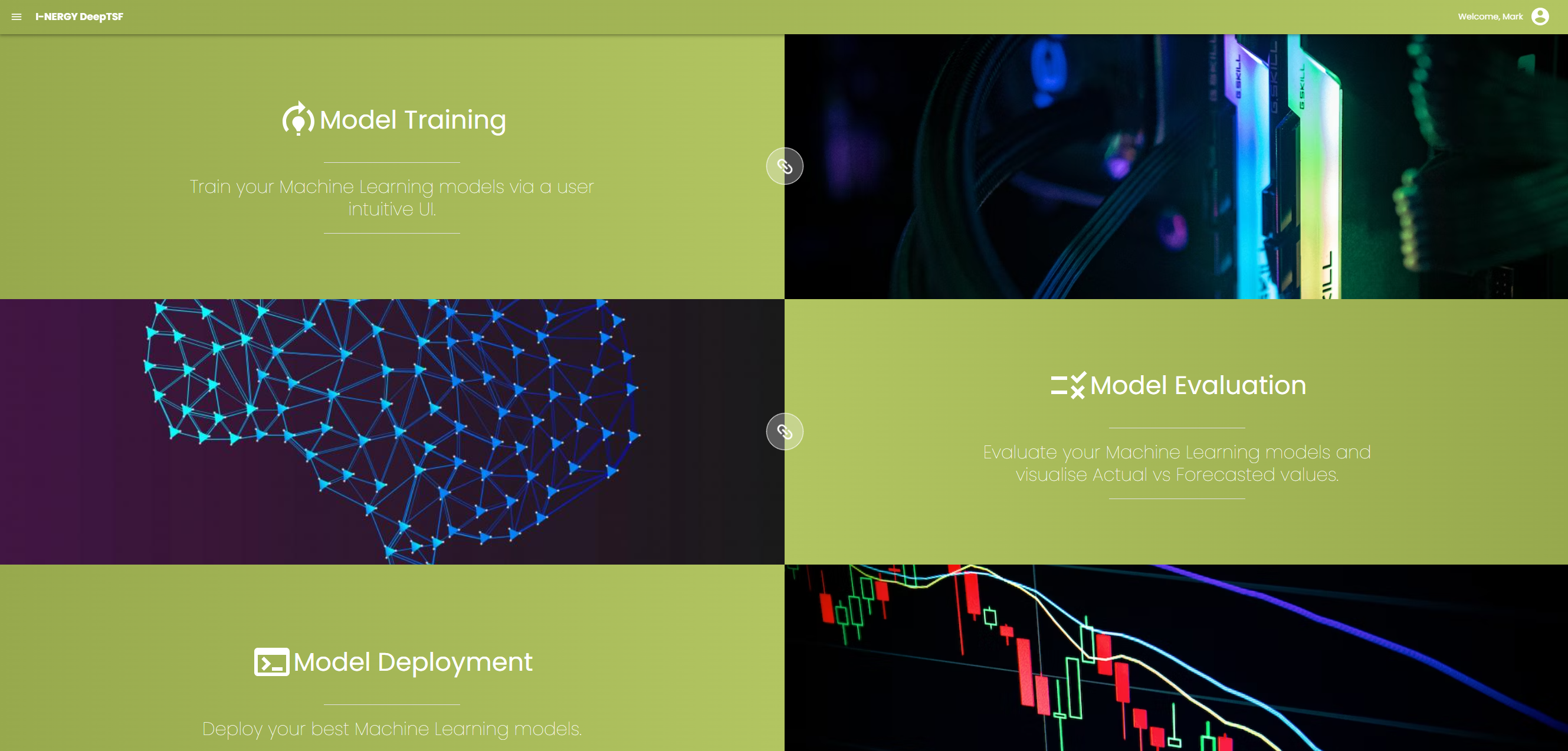}
   \caption{DeepTSF Homepage.}
   \label{fig:homepage}
\end{subfigure}
\caption{Introductory web pages of DeepTSF}
\end{figure*}

{\textbf{Sign-In and homepage}} The Sign-In page has been developed to provide the capability of logging into the dashboard, as illustrated in Fig.~\ref{fig:signin}. After signing in, the homepage displays the main capabilities of the dashboard in head titles, as illustrated in Fig.~\ref{fig:homepage}. Visually, the homepage is separated in horizontal sections, one for each one of the UI's main functionalities. 




\par{\textbf{{Codeless forecast}}
This page enables the execution of an experiment, after uploading the desired dataset in csv format and configuring key parameters of the model pre-processing, training and model evaluation steps. The provided interface has been divided in small, distinct sections, which guide the data scientist through the needed steps to sufficiently provide the required input for the execution of an experiment as illustrated in Fig.~\ref{fig:forecast}. In the Dataset Configuration section, the users can either upload their own files or choose among the already stored ones, before selecting the time series resolution and the dataset split that contains the start and end dates, which represent the dates defining the experiment execution. In the Model Training Setup section, the users can define the experiment name, choose between the available algorithms and choose between a set of hyperparameters that vary depending on the chosen algorithm. Finally, the users have to choose a forecast horizon for backtesting the model during evaluation. After filling all the required fields, the “EXECUTE” button becomes available. Upon clicking the button, an experiment with the entered configuration runs and the user is given the choice to navigate to the MLflow tracking UI to retrieve details about the execution.
\begin{figure}[H]
    \centering
    \includegraphics[width=\textwidth]{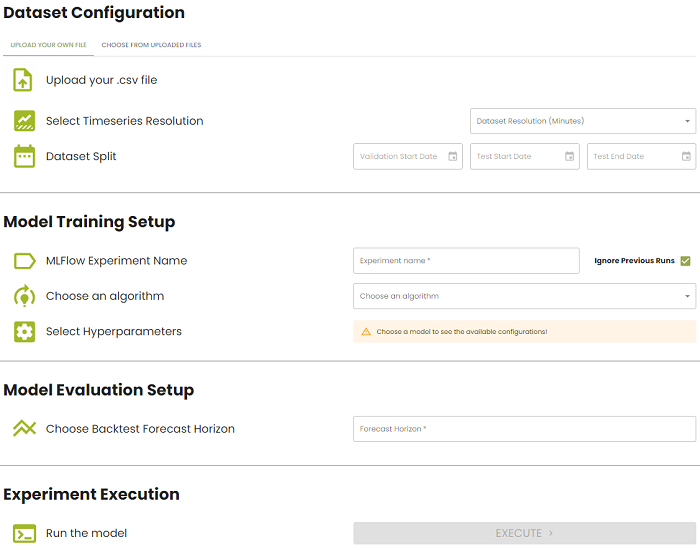}
        \caption{DeepTSF codeless forecast page.}
    \label{fig:forecast}
\end{figure}
\par{\textbf{Experiment tracking / evaluation}}
This page offers the users the capability of evaluating their experiments' results, by either their name or their id as illustrated in Fig.~\ref{fig:experiment_forecasting}. After specifying the main evaluation metric and a number of evaluation samples, the users can press the "DETAILS ON MLFLOW" button, which navigates them to the MLflow instance, where they can access a number of comprehensive experiment details. Moreover, upon clicking the "LOAD METRICS", two charts are displayed on the bottom of the page, as demonstrated in Fig.~\ref{fig:experiment_forecasting_charts}. The first chart presents the model evaluation metrics of the specified experiment, while the second one offers a visual comparison of the actual and the forecasted load series.
\begin{figure}[H]
    \centering
    \includegraphics[width=\textwidth]{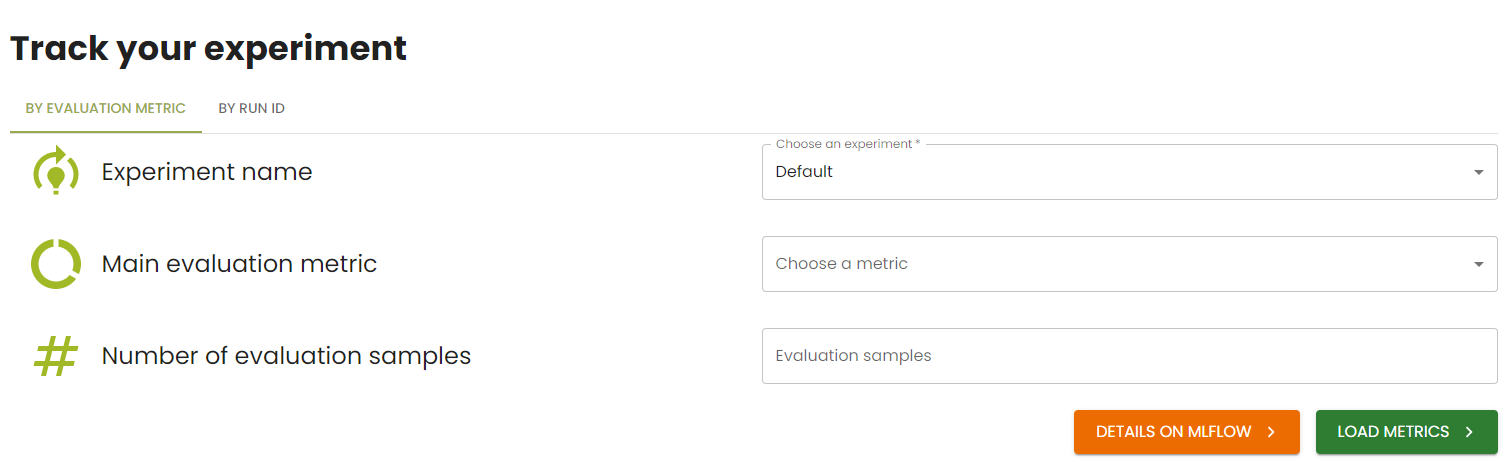}
    \caption{DeepTSF experiment tracking page.}
    \label{fig:experiment_forecasting}
\end{figure}

\begin{figure*}[htb!]
\centering
\begin{subfigure}[b]{.48\linewidth}
   \centering
   \includegraphics[height=3.2cm, width=\textwidth]{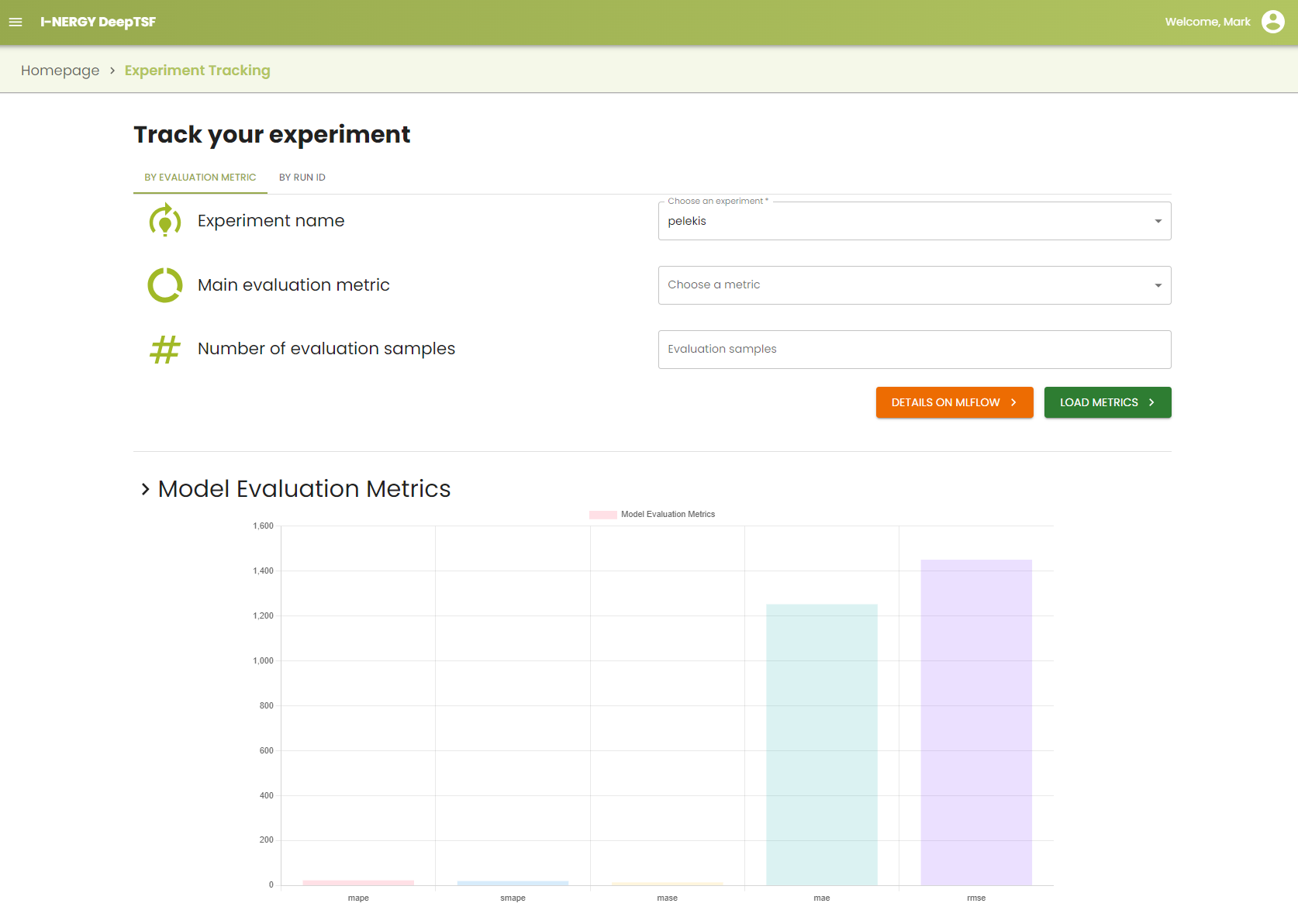}
   \caption{Evaluation metrics of the selected model for inspection.}
   \label{fig:metrics}
\end{subfigure}
\hfill
\begin{subfigure}[b]{.48\linewidth}
   \centering
   \includegraphics[height=3.2cm, width=\textwidth]{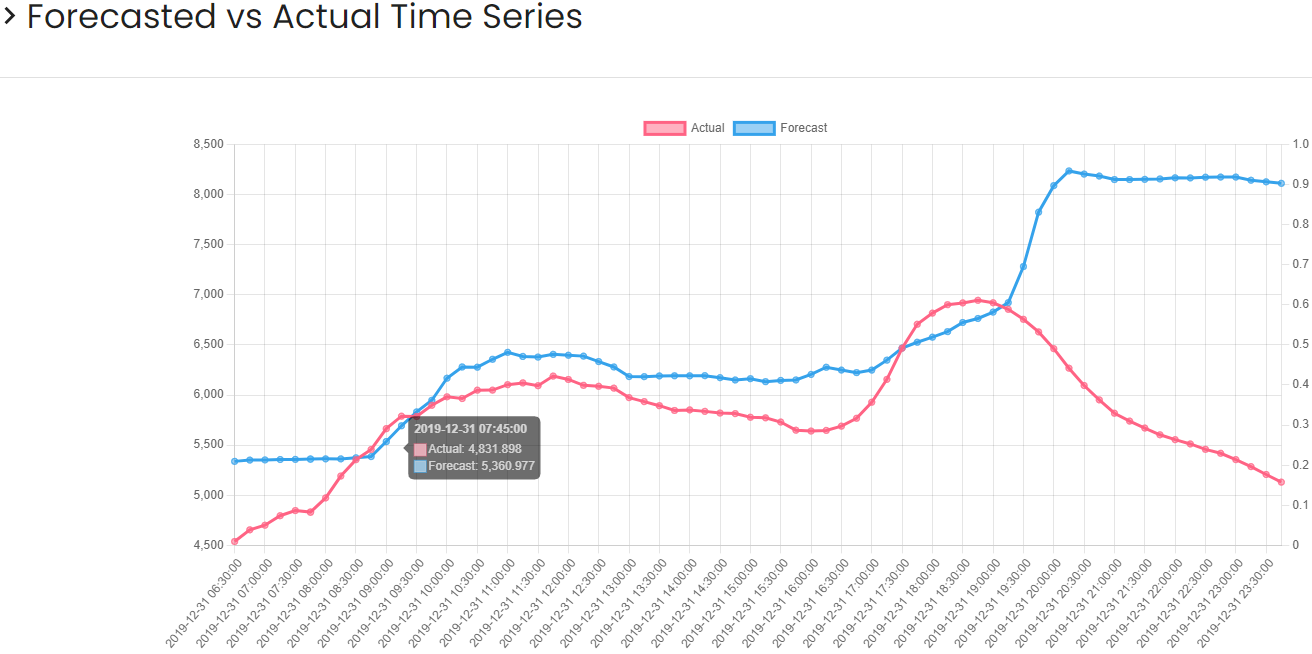}
   \caption{Line plot of the forecast versus actual time series.}
   \label{fig:actualvsforecast}
\end{subfigure}
\caption{DeepTSF experiment tracking and evaluation page results.}
\label{fig:experiment_forecasting_charts}
\end{figure*}

\par{\textbf{System monitoring}}
The system monitoring page presents a user interface that showcases real-time data pertaining to the overall memory, GPU, and CPU utilization of the deployed infrastructure, as illustrated in Fig.~\ref{fig:system_monitoring}. The displayed data are continually refreshed at a one-second interval. To prevent overloading the backend responsible for providing this information, the live demonstration is limited to one minute (60 reloads). This configuration ensures that extended periods of inactivity, such as leaving the tab open, do not strain the backend. As a result, each section includes a "Refresh Live Feed" button, enabling users to re-engage with the live monitoring if desired.
\begin{figure}[H]
    \centering
    \includegraphics[width=\textwidth]{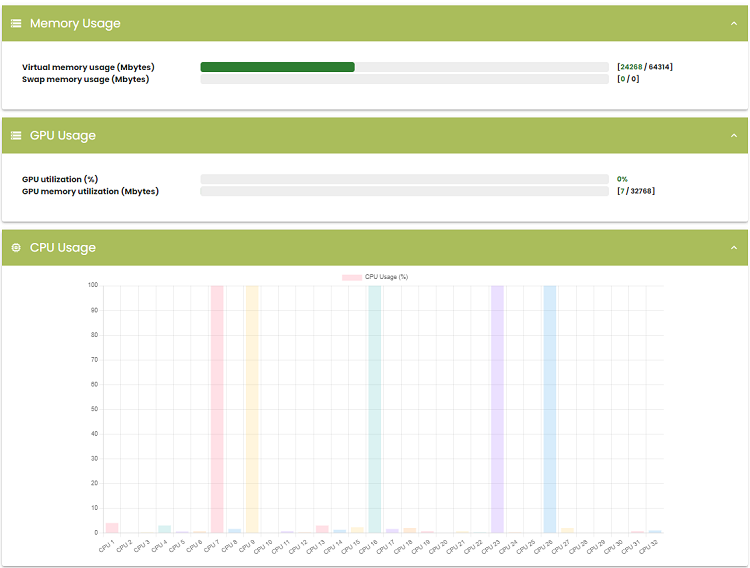}
    \caption{DeepTSF System Monitoring}
    \label{fig:system_monitoring}
\end{figure}

\subsubsection{Forecasting interface}
\label{sec:2:forecastinterface}
As above mentioned, the main ML workflow comprises 4 consecutive steps, as illustrated in Fig. \ref{fig:Experimentation_pipeline}.  The executed experiments are logged to the MLflow tracking server and can be visualized by the user using the MLflow UI that is demonstrated in Section \ref{sec:3:CLI}. The functionalities of each stage are described in more detail in the following subsections.

\begin{figure}[H]
    \centering
    \includegraphics[width=\textwidth]{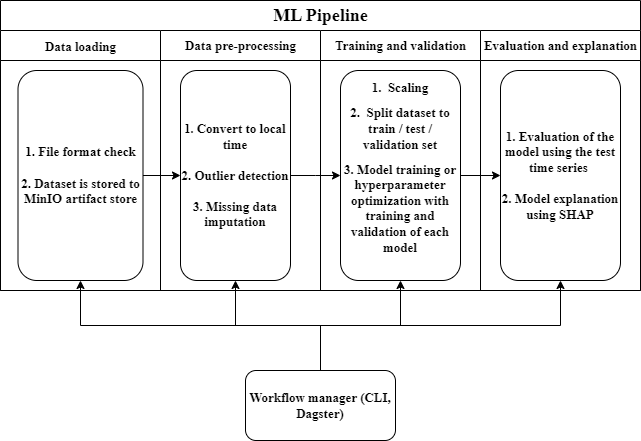}
    \caption{DeepTSF backend standard workflow}
    \label{fig:Experimentation_pipeline}
\end{figure}

\par{\textbf{Data loading}}
DeepTSF can handle univariate, multivariate, and multiple time series, optionally including external variables (covariates). Firstly, the dataset is loaded from local or online sources. Currently Deep-TSF supports csv files of the schema that is included in \ref{app:c}. In this context, the connectors that enable data ingestion vary depending on the use case and the schema of the respective data source and shall be engineered by the DeepTSF user. We provide an example of this connector, which works with MongoDB. After that, validation is performed to ensure that the files provided by the user respect the required schema. The files are saved on the MLflow tracking server so that they are available for the data pre-processing stage. 

\par{\textbf{Data pre-processing}} \label{data_preprocessing}
Then, data pre-processing is performed. Specifically, for each component of each time series, outlier detection is optionally conducted by removing values that differ more than an arbitrary number (defined by the user) of standard deviations from their monthly average, or that are zero in the case of a non-negative time series. Outliers are replaced by missing values. Subsequently, missing data may be imputed by using a weighted average of historical data and linear interpolation as proposed by \citet{Peppanen2016HandlingImputation}.

\par{\textbf{Training and validation}}
After the pre-processing stage, the data is scaled using min-max scaling, and is split into training, validation, and testing data sets. Then, the training of the model begins using only the training data set. The currently supported models are N-BEATS \citep{Oreshkin2019N-BEATS:Forecasting}, Transformer \citep{Vaswani2017AttentionNeed}, NHiTS \citep{Challu2022N-HiTS:Forecasting}, temporal convolutional networks \citep{Bai2018AnModeling}, (block) recurrent neural networks \citep{Hochreiter1997a}, temporal fusion transformers \citep{Lim2019TemporalForecasting}, LightGBM \citep{Ke}, random forest \citep{Breiman2001}, and seasonal naive as from the documentation of Darts forecasting models \citep{Darts2023DartsModels}. The latter can serve as an effective baseline depending on the seasonality of the time series \citep{Pelekis2023ADrivers}. Hyperparameter optimization can be also triggered using the Optuna library. DeepTSF supports both exhaustive and Tree-Structured Parzen Estimator-based \citep{Bergstra2011AlgorithmsOptimization} hyperparameter search. The first method tests all possible combinations of the tested hyperparameters, while the second one uses probabilistic methods to explore the combinations that result to optimal values of the user-defined loss function. Ultimately, a method based on functional analysis of variance (fANOVA) and random forests, as proposed by \citep{Hutter2014AnImportance} is used to calculate the importance of each hyperparameter during optimization. In this context, a bar graph showing the aforementioned results is produced and stored as an artifact to MLflow. 

\par{\textbf{Evaluation and explanation}}
When model training is complete, evaluation is performed through backtesting on the testing data set. Specifically, for each time series given to the function, it consecutively forecasts time series blocks of length equal to the forecast horizon of the model from the beginning until the end of the test set. This operation takes place by default with a stride equal to forecast horizon but can be changed by the user. Then, evaluation metrics are calculated using the resulting forecasted time series. The evaluation metrics that are supported are: mean absolute error (MAE), root mean squared error (RMSE), min-max and mean normalized mean squared error (NRMSE), mean absolute percentage error (MAPE), standardized mean absolute percentage error (sMAPE), and mean absolute scaled error (MASE) \citep{Hyndman2006}. In the case of multiple time series, it is possible for all evaluation sub-series to be tested leading to an average value for each one of the metrics. In this case, DeepTSF stores the results for all time series. Additionally, it is possible to analyze the output of DL and DL models using SHapley Additive exPlanations \citep{Lundberg2017APredictions}. Each SHAP coefficient indicates how much the output of the model changes, given the current value of the corresponding feature. In DeepTSF's implementation, the lags after the start of each sample are considered as the features of each model. Following that, a beeswarm plot \citep{SHAPSHAPPlot} is produced. In addition, a minimal bar graph is produced showing the average of the absolute value of the SHAP coefficients for each attribute. Finally, three force plot charts are produced, showing the exact value of its SHAP coefficients for a random sample. The above mentioned artifacts are accessible through the MLflow tracking UI.

\subsubsection{Workflow management interface} \label{workfloworchestration}
As already mentioned in Section \ref{sec:2:architecture}, the workflow management interface comprises two subcomponents: i) a CLI for triggering the stages of the pipeline, ii) a workflow orchestrator based on Dagster that takes care of model re-training and error handling issues.

\par{\textbf{CLI}} The DeepTSF CLI is a direct way of handling and triggering the workflow stages easily, automating the argument passing process and linking the execution of each script in a sequential order, passing the proper arguments from one to another. Figure \ref{fig:Experimentation_pipeline} illustrates this concept. More details on the DeepTSF CLI functionality for either the manual execution of pipeline stages or their automated sequential execution can be sought in Section \ref{sec:illustrative_examples} as well as the MLproject file that is contained in \ref{app:a}.

\par{\textbf{Dagster}} Dagster \citep{Dagster2023DagsterPipelines} acts as a workflow orchestration engine, where data processing and ML model training pipelines, are defined as jobs. Therefore, the execution of these jobs can be scheduled in fixed intervals, serving the needs of periodic training. This component interacts with the data source, extracting the data needed for the pipelines, as well as with the model registry, where the models are stored when training is completed. More specifically, the defined jobs in Dagster are in line with the flow described in section \ref{sec:2:forecastinterface}, starting with a Dagster asset responsible to load the raw data, from database using the defined resource modeling the MongoDB. The next step of the workflow consists of the pre-processing task, producing the asset that will be consumed by the following training step and (optionally) validation step. When training is complete, the produced trained model is stored to the MLflow Models registry. The last step consists of the model evaluation process. The MLflow tracking server is utilized along steps to track the ML experiment, storing information such as the calculated metrics in evaluation phase, the best parameters computed in hyperparameter tuning, and the configuration options used for the execution of pipeline. The execution of this pipeline is scheduled in order to be in sync with the periodic update of the MongoDB loading the new smart meters data. Ultimately, Dagster provides a web user interface including, among others, information regarding the defined jobs and their runs, the defined schedules, the produced assets, as well as providing the ability to configure and execute these jobs.

\subsubsection{User roles}
Access to components of DeepTSF is restricted to authorized users. Users should first be registered using the identity and access control mechanism of DeepTSF, in order to use the components of the application. Then, based on their roles, they acquire different levels of access to the application. Supported roles include data scientists, domain experts, and administrators. The "domain expert user role" has limited access to application features that are the "experiment tracking" and "system monitoring" interfaces. The "data scientist" user role has additional access to the "load forecast" feature of the main page while they are also allowed to straight access to the API of the application back-end. The MLflow experiments are available to all users. Lastly, the administrator role have access to all the features of the application.

\section{Illustrative Examples} \label{sec:illustrative_examples}
In this section we provide illustrative examples of how a data scientist can interact with DeepTSF. As already mentioned, there are two ways in which this user can interact with the application: i) through the DeepTSF UI, ii) through the DeepTSF CLI.

\subsection{A use case on DeepTSF's CLI functionality} \label{sec:3:CLI} 
In this section we provide an illustrative use case of DeepTSF's CLI functionality which is aimed for coding experts such as data scientists and machine learning engineers. Specifically, we will utilize the CLI for day-ahead short-term load forecasting (STLF) on Italy's national electricity load. Within our example, the N-BEATS deep learning model has been chosen, given its high performance on similar use cases \citep{Pelekis2022InPerformance, Pelekis2023ADrivers}.

\par The command that should be given through the terminal to DeepTSF's CLI is shown in the code snippet of Fig. \ref{fig:command}. It stores its results in the MLflow experiment named "example", and executes the entire workflow. 

\begin{figure}[H]
\begin{minted}[fontsize=\footnotesize, linenos=False, frame=lines, framesep=2mm, breaklines]{shell}
mlflow run --experiment-name example --entry-point exp_pipeline . -P from_mongo=false -P series_csv=Italy.csv -P convert_to_local_tz=false -P day_first=false -P from_mongo=false -P multiple=false -P l_interpolation=false -P resolution=60 -P rmv_outliers=true -P country=IT -P year_range=2015-2022 -P cut_date_val=20200101 -P cut_date_test=20210101 -P test_end_date=20211231 -P scale=true -P darts_model=NBEATS -P hyperparams_entrypoint=NBEATS_example -P loss_function=mape -P opt_test=true -P grid_search=false -P n_trials=100 -P device=gpu -P ignore_previous_runs=t -P forecast_horizon=24 -P m_mase=24 -P analyze_with_shap=False --env-manager=local
\end{minted}
\caption{Full command given to the DeepTSF CLI for the execution of the Italian STLF use case}
\label{fig:command}
\end{figure}
\par

The options corresponding to all steps of the pipeline can be set by the user through the CLI command. We will mainly refer to a subset of options that are relevant to our STLF use case, while the rest of them can be looked up in DeepTSF's documentation \citep{DeepTSF2023DeepTSFDocumentation}. The options that follow are presented in accordance with the step of the pipeline they refer to.

\begin{itemize}
\item \textbf{Data loading options:} Whether to take the data set from MongoDB or not (from\_mongo = false), which data set file to use (series\_csv = Italy.csv), whether the data set has dates with the day appearing before the month or not (day\_first = false), whether to get the dataset from mongo database or not (from\_mongo = false), and if the time series is multiple (multiple = false).

\item \textbf{Data pre-processing options:} If we want to convert to local timezone (convert\_to\_local\_tz = false), whether to perform imputation with the method described in \ref{data_preprocessing} or just using linear interpolation (in this case we use the former, linear\_interpolation = false), the resolution of the data set (in this case 60 minutes, resolution = 60), if the outliers will be removed (remove\_outliers = true), the country code to use for obtaining the holidays for our imputation method (country = IT), and the years of the time series to use (year\_range = 2015-2022).

\item \textbf{Training and validation options:} How the data will be split in train-validation-test sets (in this case we use years 2015-2019 as a train set, 2020 as a validation set, and 2021 as a test set. To do that we set cut\_date\_val = 20200101, cut\_date\_test = 20210101, and test\_end\_date = 20211231), whether to scale the data set or not (scale = true), which model type to use (darts\_model = NBEATS), from which entry point to get the model parameters (hyperparams\_entrypoint = NBEATS\_example, see \ref{app:d} about the way we provide the parameters to DeepTSF), which metric to use as an objective function (loss\_function = MAPE) throughout the TPE optimization process, whether we desire to perform hyperparameter optimization (opt\_test = true), whether to perform an exhaustive search or use the TPE method (in this case the latter, grid\_search = false), the number of trials of hyperparameter search to execute (n\_trials = 100), whether to train the model using the GPU, or just the cpu(device = gpu), and whether to ignore previous stages of the workflow that have been previously executed with the same arguments (ignore\_previous\_runs = t) or use them as a starting point instead.

\item \textbf{Evaluation and explanation options:} The forecast horizon to be used during backtesting (in this case 24 hours, forecast\_horizon = 24), the forecast horizon of the naive method used in MASE (m\_mase = 24), and whether to perform an analysis using SHAP or not (analyze\_with\_shap = False).

\end{itemize}
All other options not present in the command take their default values. 
\par After the command is run, the pipeline is executed. The various stages appear as separate MLflow child runs of the main run in the MLflow UI (Fig. \ref{fig:mlflow_ui}). The user can visualize the results related to each stage by clicking on the corresponding run. The hyperparameter optimization stage's page is illustrated in Fig. \ref{fig:optuna_stage}.

 \begin{figure}[H]
    \centering
    \includegraphics[width=\textwidth]{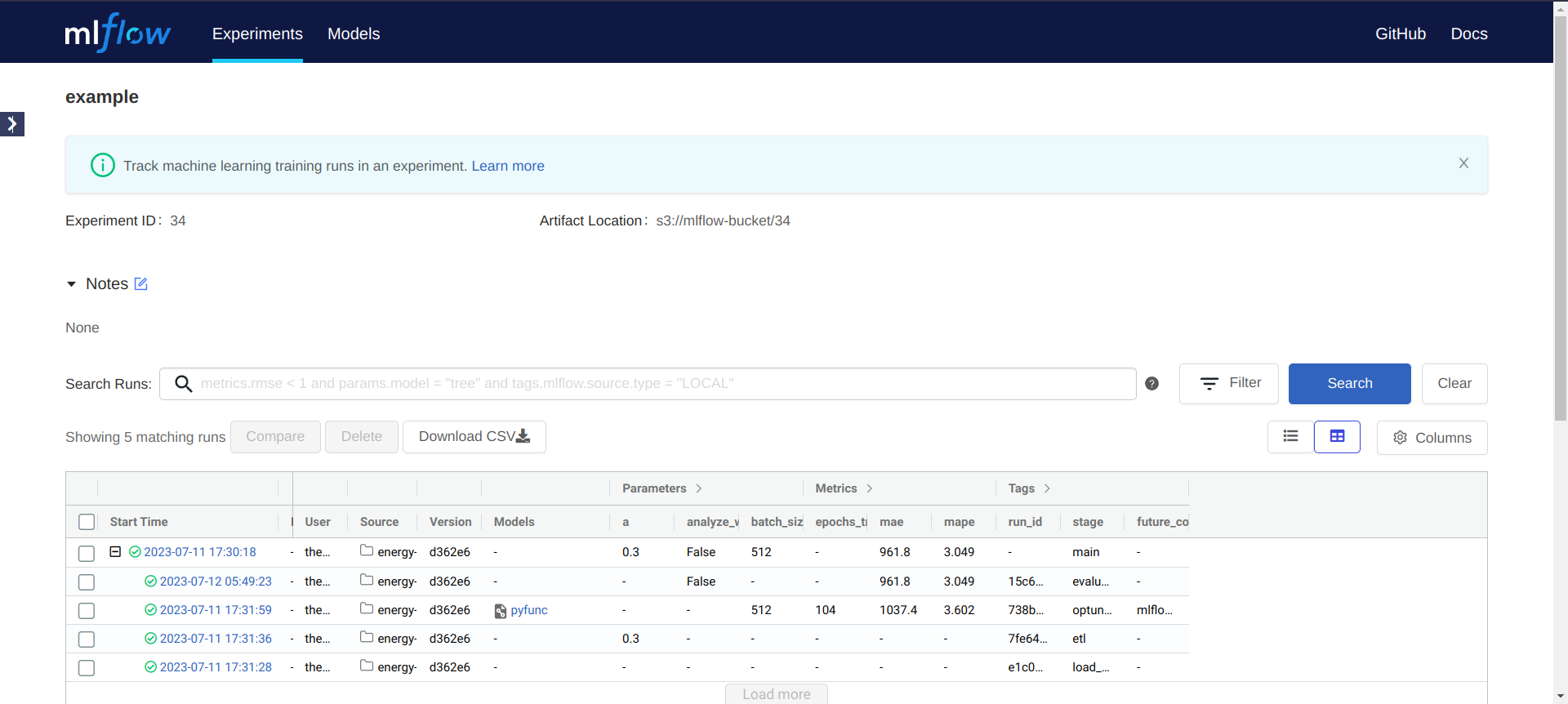}
    \caption{MLflow user interface example}
    \label{fig:mlflow_ui}
\end{figure}

\begin{figure}[H]
    \centering
    \includegraphics[width=\textwidth]{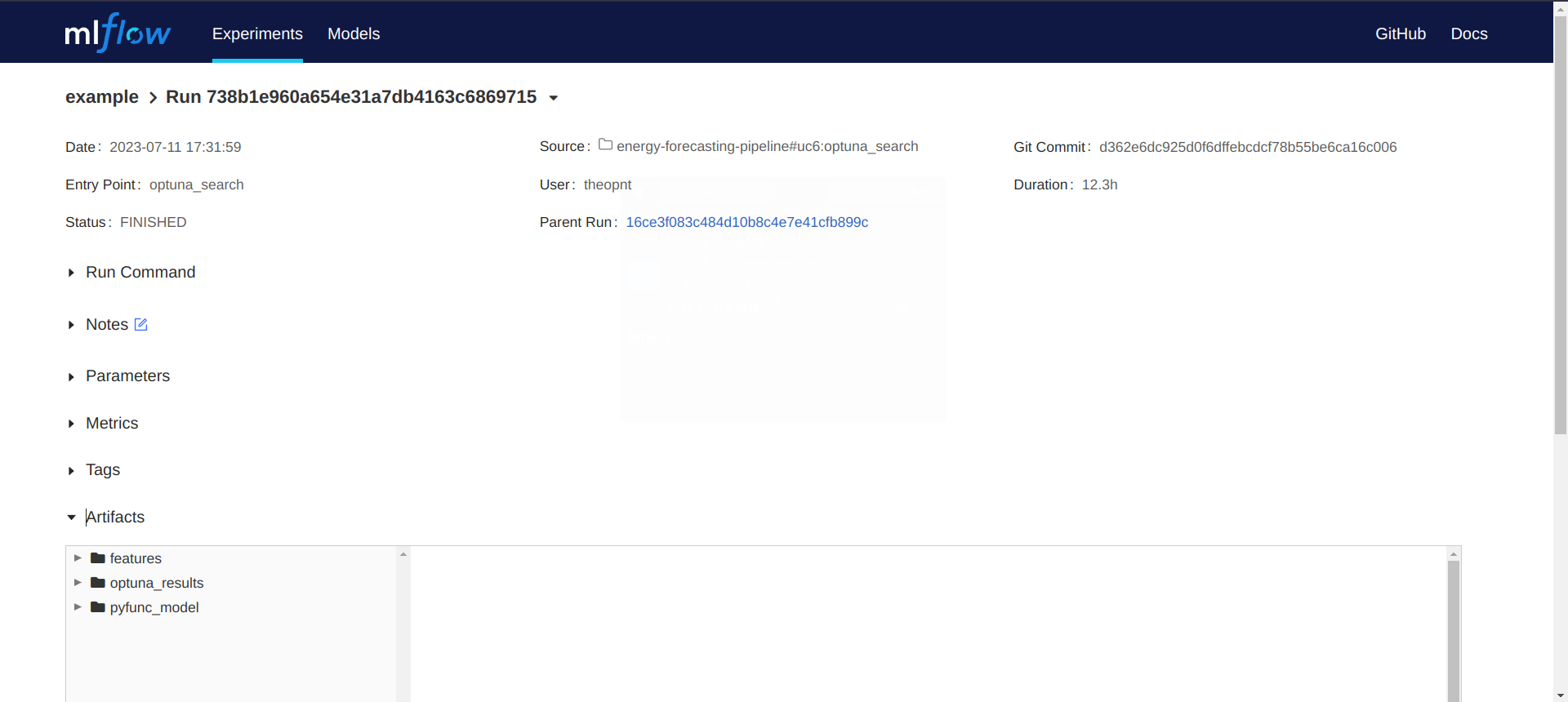}
    \caption{Hyperparameter optimization stage}
    \label{fig:optuna_stage}
\end{figure}

\par 
Deep diving into the results of this study, the data loading stage saved the data set on the MLflow tracking server after validation of the time series was performed. The data pre-processing stage found no outliers to remove and imputed 48 missing values. The resulting time series can be visualized as illustrated in Fig. \ref{fig:imputed_series}.

\begin{figure}[H]
    \centering
    \includegraphics[width=\textwidth]{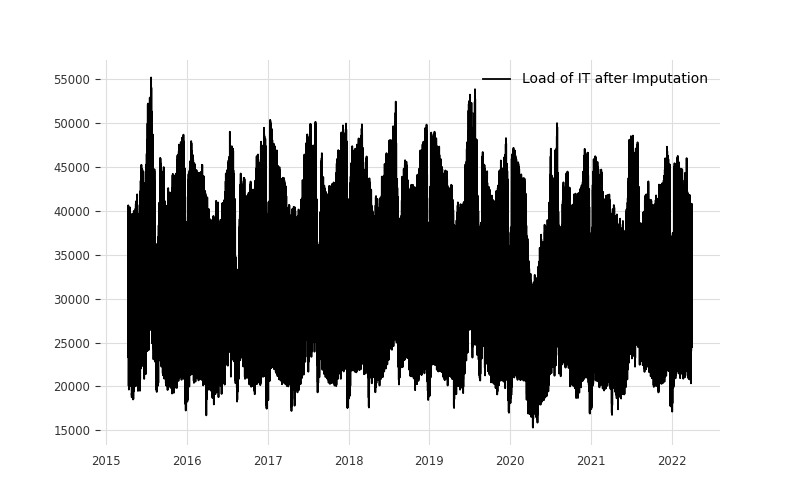}
    \caption{Time series line plot after the imputation of missing values by DeepTSF}
    \label{fig:imputed_series}
\end{figure}

\par
The hyperparameter optimization stage was executed for 100 trials, and the best model was saved alongside its parameters. The MAPE objective function was calculated on the validation set (2020). The MAPE value of each trial can be seen in Fig. \ref{fig:optuna_history}. More detailed results about each trial can be found in the product csv file "NBEATS\_example.csv", a sample of which is provided in Table \ref{tab:optuna_results_csv}. Each trial's hyperparameters, as well as all the corresponding evaluation metrics are included there.

\begin{figure}[H]
    \centering
    \includegraphics[width=\textwidth]{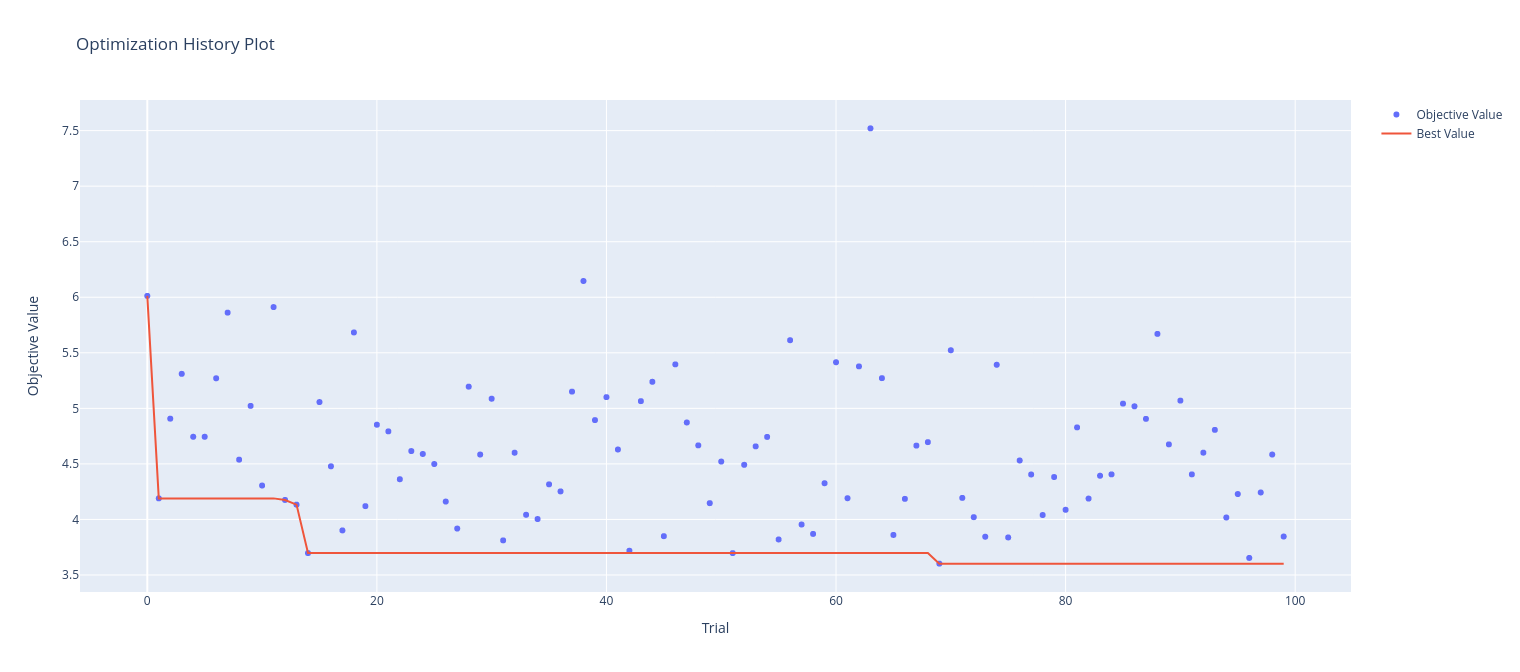}
    \caption{The values of the objective function (MAPE) for each hyperparameter optimization trial}
    \label{fig:optuna_history}
\end{figure}

\begin{table}[!ht]
    \centering
    \caption{Detailed results of all hyperparameter optimization trials}
    \label{tab:optuna_results_csv}
    \begin{tabular}{|l|l|l|l|l|l|l|l|l|l|l|l|l|l|l|l|l|l|l|l|l|l|}
    \hline
        number & value & datetime\_start & ... & batch\_size & num\_blocks & ... \\ \hline
        0 & 6.01 & 2023-07-11 16:32:05 & ... & 1536 & 6 & ... \\ \hline
        1 & 4.19 & 2023-07-11 16:35:31 & ... & 1280 & 6 & ... \\ \hline
        2 & 4.91 & 2023-07-11 16:42:50 & ... & 512 & 6 & ... \\ \hline
        3 & 5.31 & 2023-07-11 16:49:13 & ... & 2048 & 9 & ... \\ \hline
        4 & 4.74 & 2023-07-11 16:56:14 & ... & 2048 & 4 & ... \\ \hline
        5 & 4.74 & 2023-07-11 17:01:40 & ... & 1024 & 6 & ... \\ \hline
        6 & 5.27 & 2023-07-11 17:08:33 & ... & 2048 & 1 & ... \\ \hline
        7 & 5.86 & 2023-07-11 17:14:40 & ... & 512 & 8 & ... \\ \hline
        8 & 4.54 & 2023-07-11 17:26:18 & ... & 256 & 7 & ... \\ \hline
    \end{tabular}
\end{table}
\par Finally, the model is evaluated at the evaluation stage. The model's prediction of the test set, along with the actual values of the series can be visualized in Fig. \ref{fig:prediction}. The resulting values on the test set of all the metrics DeepTSF supports are saved as MLflow metrics (see Fig. \ref{fig:metrics_eval}).

\begin{figure}[H]
    \centering
    \includegraphics[width=\textwidth]{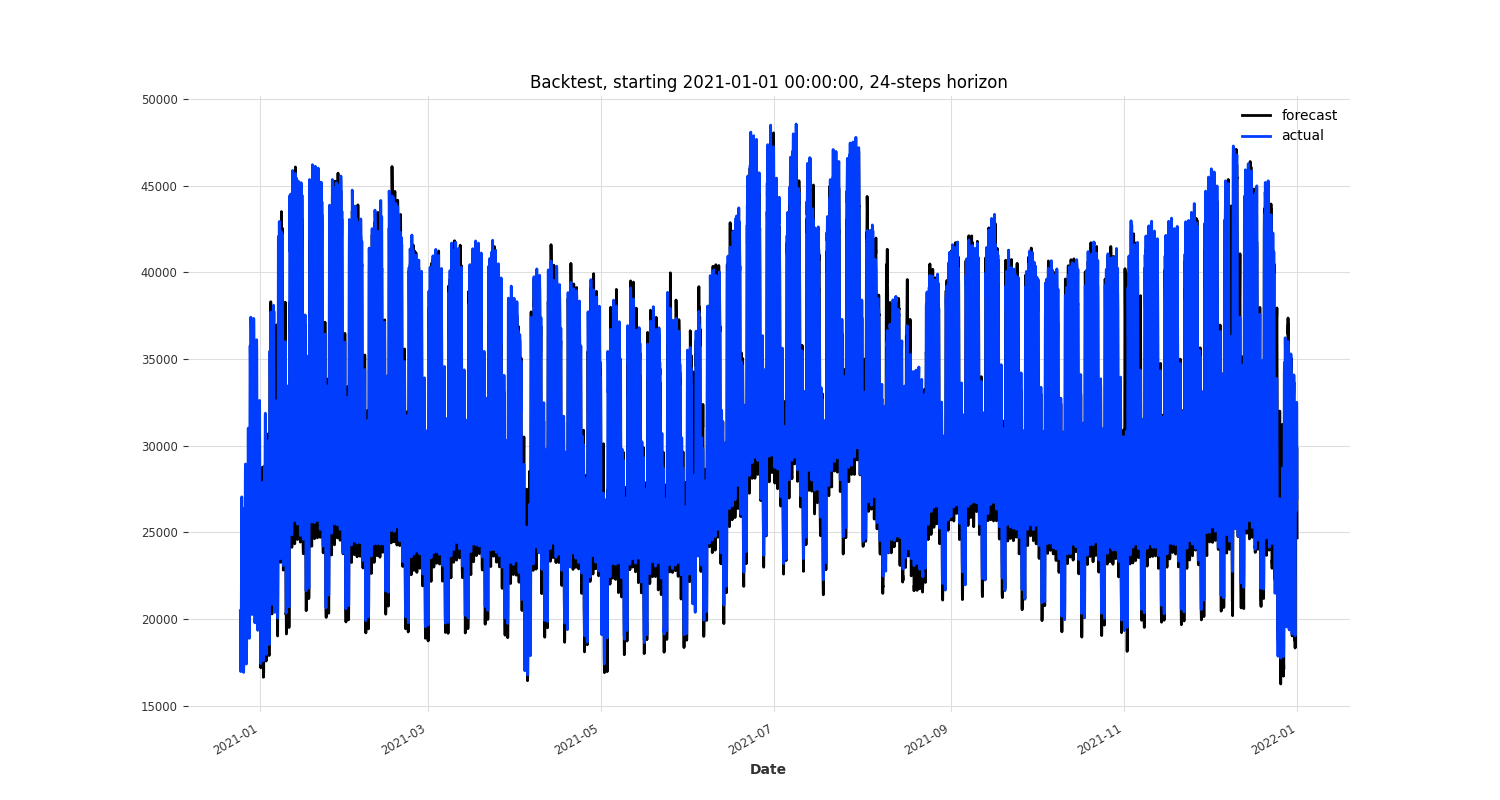}
    \caption{Forecast versus actual plot produced by the model evaluation stage and stored as artifact to MLflow/MinIO}
    \label{fig:prediction}
\end{figure}

\begin{figure}[H]
    \centering
    \includegraphics[width=\textwidth]{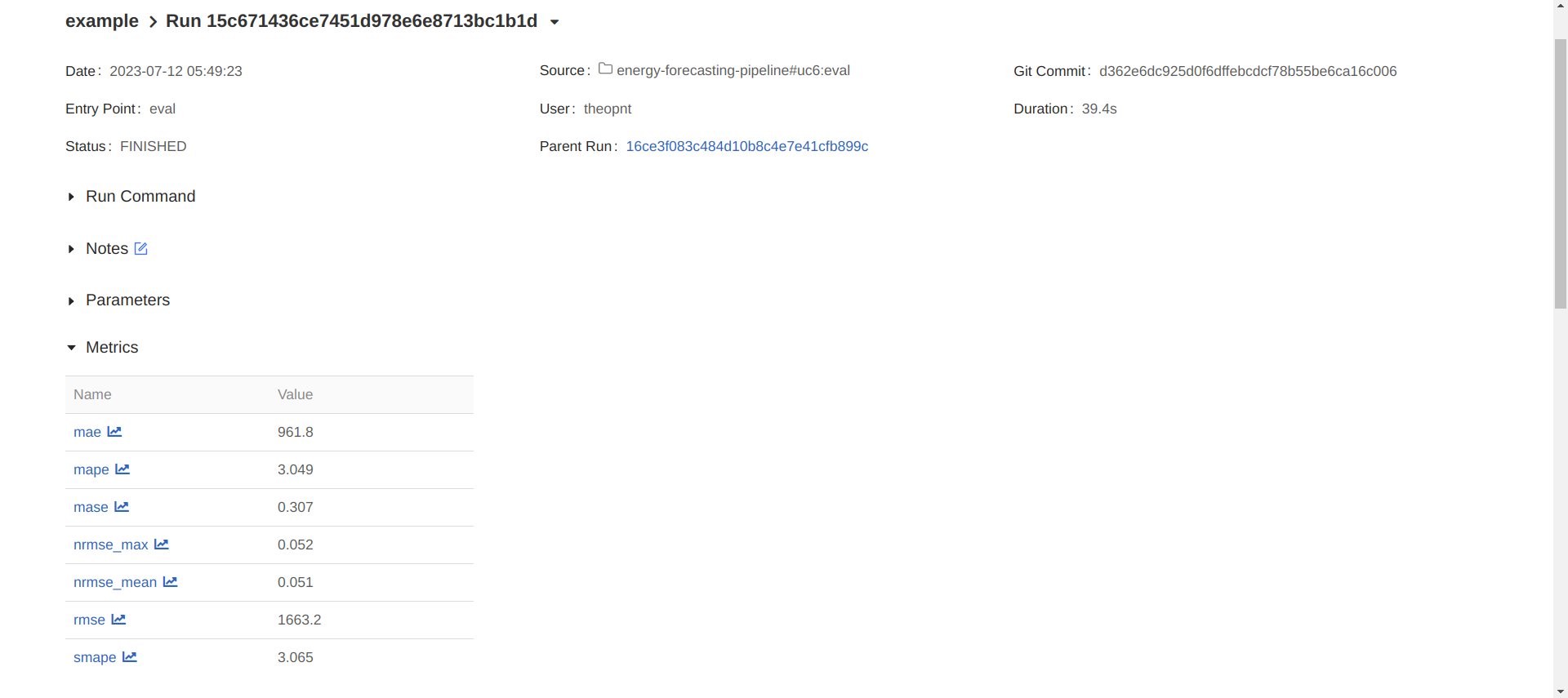}
    \caption{Performance metrics of the forecasting model on the test set as calculated during the evaluation stage}
    \label{fig:metrics_eval}
\end{figure}
    Further information on the output files of DeepTSF can be found in the documentation of the software \citep{DeepTSF2023DeepTSFDocumentation}.

\subsection{Codeless model training for data scientists via DeepTSF's UI} \label{sec:3:codeless} 
For a data scientist that desires a fully codeless experience, DeepTSF offers the capability of training and evaluating models through the UI as already mentioned. To this end, the user needs to follow the steps below\footnote{This section briefly describes the steps that need to be followed by the user to do that by referring to descriptions and figures from Section 2, in order to avoid content repetition.}:

\begin{enumerate}
    \item Use the Sign-In page and provide their credentials (see Fig. \ref{fig:signin})
    \item Navigate through the homepage, open the side dashboard and select the "Load Forecast" option (see Fig. \ref{fig:homepage})
    \item Follow the process described within the "Codeless forecast" paragraph of Section \ref{sec:2:ui} (see Fig.~\ref{fig:forecast})
    \item To evaluate the model in detail, the MLflow UI can be visited as described in the previous section. This can be done by clicking the "Visit MLflow server" button which becomes available after the execution.
    \item To graphically evaluate the model at a higher-level, the user needs to follow the process described within the "Experiment tracking and evaluation" paragraph of Section \ref{sec:2:ui} (see Fig.~\ref{fig:experiment_forecasting} and Fig.~\ref{fig:experiment_forecasting_charts})
\end{enumerate}

\section{Impact} \label{impact}







Since DeepTSF is open-source, it can serve as a reference production tool for stakeholders such as data scientists, and domain experts that need to develop, optimize, evaluate, serve and monitor ML and DL models for time series forecasting. DeepTSF makes MLOps for time series forecasting easy for everyone by unifying the ML lifecycle development and monitoring. This is achieved through a user interface that guarantees a user-friendly and codeless experience which is its main contribution to the industrial time series forecasting domain. Additionally, CLI capabilities are also available for data experts and ML engineers that require custom model training procedures and flexibility, given a specific degree of coding expertise. 

Thanks to its design, DeepTSF enables the codeless development, initialization, and seamless monitoring of ML experiments. In this fashion, the mass execution of ML experiments is facilitated. Therefore, researchers can easily reproduce and evaluate an abundance of research results related to various ML and DL models and hyperaparameter setups. In this fashion, DeepTSF reinforces the ability for research contributions within the time series forecasting domain. Within industrial environments, DeepTSF can accelerate the processes for the deployment of accurate time series forecasting models in production as domain experts are allowed to directly monitor the model development process via intuitive user interfaces that match their domain knowledge even if they possess limited modeling skills.

DeepTSF is already used in the EU H2020 research project I-NERGY \citep{Karakolis2022ARTIFICIALPROJECT} and it was developed as a solution to enable automated energy time series forecasting for both generation and consumption time series for multiple pilot sites and use cases within the project. The service was meant to serve both data scientists coming from the technical partners of the project and energy experts coming from the pilot partners and coordinate their every day time series forecasting tasks in an efficient and highly collaborative fashion. 

\section{Conclusions and future work} \label{conclusion}

\subsection{Conclusions}
In this paper, we introduced DeepTSF, a comprehensive machine learning operations (MLOps) framework designed to revolutionize time series forecasting with codeless machine learning (ML) capabilities. DeepTSF automates several aspects of the ML lifecycle, making it an ideal tool for data scientists and MLops engineers engaged in ML and DL-based forecasting.

By incorporating cutting-edge ML and DL algorithms, DeepTSF empowers users with a robust and user-friendly solution for precise and efficient time series forecasting. Leveraging the power of Python's extensive scientific libraries and frameworks, DeepTSF seamlessly integrates into existing data analysis workflows, enhancing productivity and compatibility. DeepTSF also includes a front-end that caters to various high-level stakeholders and end-users in the time series modeling domain. Through insightful visualizations and evaluation metrics, these stakeholders gain a comprehensive understanding of the forecasting models developed by engineers at a lower level. Security is also top priority for DeepTSF, and it ensures protection by implementing effective identity management and access authorization.

DeepTSF is versatile and capable of handling all types of forecasting tasks. In real-world applications, currently DeepTSF has already proven its efficacy in DL-based short-term electricity load forecasting use cases with significant added value in the electrical power and energy systems sector.

\subsection{Future work}
With respect to future work we plan to further extend DeepTSF's workflow orchestration interface so that it can seamlessly handle the data and ML pipelines through DAGs, therefore fully replacing the capabilities that are currently supported by the CLI.

Additionally, regarding model serving, it is envisaged that DeepTSF will also integrate with other model serving mechanisms such as Seldon \citep{Seldon2023SeldonDeployment}, BentoML \citep{BentoML2023BentoML:Applications} and interoperability frameworks such as ONNX \citep{ONNX2023ONNXHome}, as MLflow's model serving features are currently on experimental stages. 

Last but not least, it is of crucial importance to make DeepTSF solution agnostic in terms of database technology, and data formatting standards, aiming to provide flexibility with respect to the integration with other user-preferred technology stacks and requirements. In this context, the data ingestion mechanism of DeepTSF is planned to be extended to support the integration with various data sources and database technologies, primarily focusing on time series databases such as InfluxDB \citep{Ahmad2017Hands-OnInfluxDB} and Timescale \citep{Timescale2023TimescaleTimescale}. Furthermore, interoperability frameworks will be utilized to facilitate different data schema standards.

\section*{Current code version}
\label{current_code}

Table \ref{tab:code_metadata} provides information about the current code github repository, documentation, and versions.

\begin{table}[H]
\begin{tabular}{|l|p{6.5cm}|p{6.5cm}|}
\hline
\textbf{Nr.} & \textbf{Code metadata description} & \textbf{Code metadata value} \\
\hline
C1 & Current code version & 0.0.9 \\
\hline
C2 & Permanent link to code/repository used for this code version & \url{https://github.com/I-NERGY/DeepTSF} \\
\hline
C3 & Code Ocean compute capsule & - \\
\hline
C4 & Legal Code License   & Attribution-NonCommercial 4.0 International \\
\hline
C5 & Code versioning system used & git\\
\hline
C6 & Software code languages, tools, and services used & Python, Javascript, Docker \\
\hline
C7 & Compilation requirements, operating environments \& dependencies & Docker for MLflow tracking server (\url{https://github.com/I-NERGY/mlflow-tracking-server/blob/master/docker-compose.yml}) and MLflow client (\url{https://github.com/I-NERGY/DeepTSF/blob/master/docker-compose.yml}), Miniconda3 for DeepTSF client (\url{https://github.com/I-NERGY/DeepTSF/blob/master/conda.yaml}) \\
\hline
C8 & If available Link to developer documentation/manual & \url{https://github.com/I-NERGY/DeepTSF/blob/master/README.md} \\
\hline
C9 & Support email for questions & spelekis@epu.ntua.gr \\
\hline
\end{tabular}
\caption{Code metadata}
\label{tab:code_metadata}
\end{table}

\section*{Acknowledgment}
This work has been funded by the European Union’s Horizon 2020 research and innovation programme under the I-NERGY project, grant agreement No. 101016508. Additionally, the HPC resources utilized for training and optimizing the required machine learning models in this study have been provided by the EGI-ACE project, which also receives funding from the European Union’s Horizon 2020 research and innovation programme under grant agreement No. 101017567.








\bibliographystyle{cas-model2-names}

\bibliography{references}


\onecolumn
\appendix
\appendix

\section{MLproject file} \label{app:a}

The MLproject file describes the way the ML pipeline is executed. It is in YAML format, and it defines the name of the MLflow project (in this case DeepTSF\_workflow), the file to use to build the environment the user desires to work with (in this case conda.yaml), and the entry points of our project. 
\par Each entry point corresponds to a specific stage of the pipeline describing the respective python command alongside its parameters. In this case, each entry point runs the main python file for the stage it corresponds to, and passes the parameters to the file using the Click library \citep{Click2023ClickDocumentation}. The file is shown in the next pages:   

\begin{minted}[fontsize=\footnotesize, linenos=True, frame=lines, framesep=2mm, breaklines]{yaml}
name: DeepTSF_workflow

conda_env: conda.yaml

entry_points:

  load_raw_data:
    parameters:
      series_csv: {type: str, default: series.csv}
      series_uri: {type: str, default: online_artifact}
      past_covs_csv: {type: str, default: None}
      past_covs_uri: {type: str, default: None}
      future_covs_csv: {type: str, default: None}
      future_covs_uri: {type: str, default: None}
      day_first: {type: str, default: "true"}
      multiple: {type: str, default: "false"}
      resolution: {type: str, default: 15}
      from_mongo: {type: str, default: "false"}
      mongo_name: {type: str, default: "rdn_load_data"}

    command: |
      python load_raw_data.py --series-csv {series_csv} --series-uri {series_uri} --day-first {day_first} --multiple {multiple} --resolution {resolution} --from-mongo {from_mongo} --mongo-name {mongo_name} --past-covs-csv {past_covs_csv} --past-covs-uri {past_covs_uri} --future-covs-csv {future_covs_csv} --future-covs-uri {future_covs_uri} 


  etl:
    parameters:
      series_csv: {type: str, default: series.csv}
      series_uri: {type: str, default: mlflow_artifact_uri}
      resolution: {type: str, default: 15}
      year_range: {type: str, default: None}
      time_covs: {type: str, default: "false"}
      day_first: {type: str, default: "true"}
      country: {type: str, default: "PT"}
      std_dev: {type: str, default: 4.5}
      max_thr: {type: str, default: -1}
      a: {type: str, default: 0.3}
      wncutoff: {type: str, default: 0.000694}
      ycutoff: {type: str, default: 3}
      ydcutoff: {type: str, default: 30}
      multiple: {type: str, default: "false"}
      l_interpolation: {type: str, default : "false"}
      rmv_outliers: {type: str, default: "true"}
      convert_to_local_tz: {type: str, default: "true"}
      ts_used_id: {type: str, default: "None"}
      infered_resolution_series: {type: str, default: "15"}
      min_non_nan_interval: {type: str, default: "24"}
      cut_date_val: {type: str, default: 20200101}
      infered_resolution_past: {type: str, default: "15"} 
      past_covs_csv: {type: str, default: "None"} 
      past_covs_uri: {type: str, default: "None"}
      infered_resolution_future: {type: str, default: "15"} 
      future_covs_csv: {type: str, default: "None"} 
      future_covs_uri: {type: str, default: "None"} 
    command: |
      python etl.py --series-csv {series_csv} --series-uri {series_uri} --resolution {resolution} --year-range {year_range} --time-covs {time_covs} --day-first {day_first} --country {country} --std-dev {std_dev} --max-thr {max_thr} --a {a} --wncutoff {wncutoff} --ycutoff {ycutoff} --ydcutoff {ydcutoff} --multiple {multiple} --l-interpolation {l_interpolation} --rmv-outliers {rmv_outliers} --convert-to-local-tz {convert_to_local_tz} --ts-used-id {ts_used_id} --infered-resolution-series {infered_resolution_series} --min-non-nan-interval {min_non_nan_interval} --cut-date-val {cut_date_val} --past-covs-csv {past_covs_csv} --past-covs-uri {past_covs_uri} --future-covs-csv {future_covs_csv} --future-covs-uri {future_covs_uri} --infered-resolution-past {infered_resolution_past}  --past-covs-csv {past_covs_csv} --past-covs-uri {past_covs_uri} --infered-resolution-future {infered_resolution_future} --future-covs-csv {future_covs_csv} --future-covs-uri {future_covs_uri}

  train:
    parameters:
      series_csv: {type: str, default: series.csv}
      series_uri: {type: str, default: mlflow_artifact_uri}
      future_covs_csv: {type: str, default: None}
      future_covs_uri: {type: str, default: mlflow_artifact_uri}
      past_covs_csv: {type: str, default: None}
      past_covs_uri: {type: str, default: mlflow_artifact_uri}
      cut_date_val: {type: str, default: 20200101}
      cut_date_test: {type: str, default: 20210101}
      test_end_date: {type: str, default: None}
      darts_model: {type: str, default: RNN}
      device: {type: str, default: gpu}
      hyperparams_entrypoint: {type: str, default: LSTM1}
      scale: {type: str, default: "true"}
      scale_covs: {type: str, default: "true"}
      multiple: {type: str, default: "false"}
      training_dict: {type: str, default: "None"}
      num_workers: {type: str, default: 4}
      day_first: {type: str, default: "true"}
      resolution: {type: str, default: 15}

    command: |
      python ../training.py --series-csv {series_csv} --series-uri {series_uri} --future-covs-csv {future_covs_csv} --future-covs-uri {future_covs_uri} --past-covs-csv {past_covs_csv}  --past-covs-uri {past_covs_uri} --cut-date-val {cut_date_val} --cut-date-test {cut_date_test} --test-end-date {test_end_date} --darts-model {darts_model} --device {device} --hyperparams-entrypoint {hyperparams_entrypoint} --scale {scale} --scale-covs {scale_covs} --multiple {multiple} --training-dict {training_dict} --cut-date-val {cut_date_val} --num-workers {num_workers} --day-first {day_first} --resolution {resolution}

  eval:
    parameters:
      mode: {type: str, default: remote}
      series_uri: {type: str, default: mlflow_artifact_uri}
      future_covs_uri: {type: str, default: mlflow_artifact_uri}
      past_covs_uri: {type: str, default: mlflow_artifact_uri}
      scaler_uri: {type: str, default: mlflow_artifact_uri}
      cut_date_test: {type: str, default: 20210101}
      test_end_date: {type: str, default: None}
      model_uri: {type: str, default: mlflow_artifact_uri}
      model_type: {type: str, default: pl}
      forecast_horizon: {type: str, default: 96}
      stride: {type: str, default: None}
      retrain: {type: str, default: "false"}
      input_chunk_length: {type: str, default: None}
      output_chunk_length: {type: str, default: None}
      size: {type: str, default: 10}
      analyze_with_shap: {type: str, default: "false"}
      multiple: {type: str, default: "false"}
      eval_series: {type: str, default: "Portugal"}
      cut_date_val: {type: str, default: 20210101}
      day_first: {type: str, default: "true"}
      resolution: {type: str, default: 15}
      eval_method: {type: str, default: "ts_ID"}
      evaluate_all_ts: {type: str, default: "false"}
      m_mase: {type: str, default: "1"}
      num_samples: {type: str, default: "1"}

    command: |
      python ../evaluate_forecasts.py --mode {mode} --series-uri {series_uri} --future-covs-uri {future_covs_uri} --model-type {model_type} --past-covs-uri {past_covs_uri} --scaler-uri {scaler_uri} --cut-date-test {cut_date_test} --test-end-date {test_end_date} --model-uri {model_uri} --forecast-horizon {forecast_horizon} --stride {stride} --retrain {retrain} --input-chunk-length {input_chunk_length} --output-chunk-length {output_chunk_length} --size {size} --analyze-with-shap {analyze_with_shap} --multiple {multiple} --eval-series {eval_series} --cut-date-val {cut_date_val} --day-first {day_first} --resolution {resolution} --eval-method {eval_method} --evaluate-all-ts {evaluate_all_ts} --m-mase {m_mase} --num-samples {num_samples}


  optuna_search:
    parameters:
      series_csv: {type: str, default: series.csv}
      series_uri: {type: str, default: mlflow_artifact_uri}
      future_covs_csv: {type: str, default: None}
      future_covs_uri: {type: str, default: mlflow_artifact_uri}
      past_covs_csv: {type: str, default: None}
      past_covs_uri: {type: str, default: mlflow_artifact_uri}
      resolution: {type: str, default: 15}
      year_range: {type: str, default: None}
      darts_model: {type: str, default: RNN}
      hyperparams_entrypoint: {type: str, default: LSTM1}
      cut_date_val: {type: str, default: 20180101}
      cut_date_test: {type: str, default: 20190101}
      test_end_date: {type: str, default: None}
      device: {type: str, default: gpu}
      forecast_horizon: {type: str, default: 96}
      stride: {type: str, default: None}
      retrain: {type: str, default: false}
      scale: {type: str, default: "true"}
      scale_covs: {type: str, default: "true"}
      multiple: {type: str, default: "false"}
      eval_series: {type: str, default: "Portugal"}
      n_trials: {type: str, default: 100}
      num_workers: {type: str, default: 4}
      day_first: {type: str, default: "true"}
      eval_method: {type: str, default: "ts_ID"}
      loss_function: {type: str, default: "mape"}
      evaluate_all_ts: {type: str, default: "false"}
      grid_search: {type: str, default: "false"}
      num_samples: {type: str, default: "1"}

    command: |
      python ../optuna_search.py --series-csv {series_csv} --series-uri {series_uri} --future-covs-csv {future_covs_csv} --future-covs-uri {future_covs_uri} --past-covs-csv {past_covs_csv}  --past-covs-uri {past_covs_uri} --resolution {resolution} --year-range {year_range} --darts-model {darts_model} --hyperparams-entrypoint {hyperparams_entrypoint} --cut-date-val {cut_date_val} --cut-date-test {cut_date_test} --test-end-date {test_end_date} --device {device} --forecast-horizon {forecast_horizon} --stride {stride} --retrain {retrain} --scale {scale} --scale-covs {scale_covs} --multiple {multiple} --eval-series {eval_series} --n-trials {n_trials} --num-workers {num_workers} --day-first {day_first} --eval-method {eval_method} --loss-function {loss_function} --evaluate-all-ts {evaluate_all_ts} --grid-search {grid_search} --num-samples {num_samples}



  exp_pipeline:
    parameters:
      series_csv: {type: str, default: series.csv}
      series_uri: {type: str, default: online_artifact}
      past_covs_csv: {type: str, default: None}
      past_covs_uri: {type: str, default: None}
      future_covs_csv: {type: str, default: None}
      future_covs_uri: {type: str, default: None}
      resolution: {type: str, default: 15}
      year_range: {type: str, default: None}
      time_covs: {type: str, default: "false"}
      hyperparams_entrypoint: {type: str, default: LSTM1}
      cut_date_val: {type: str, default: 20180101}
      cut_date_test: {type: str, default: 20190101}
      test_end_date: {type: str, default: None}
      darts_model: {type: str, default: RNN}
      device: {type: str, default: gpu}
      forecast_horizon: {type: str, default: 96}
      stride: {type: str, default: None}
      retrain: {type: str, default: false}
      ignore_previous_runs: {type: str, default: "true"}
      scale: {type: str, default: "true"}
      scale_covs: {type: str, default: "true"}
      day_first: {type: str, default: "true"}
      country: {type: str, default: "PT"}
      std_dev: {type: str, default: 4.5}
      max_thr: {type: str, default: -1}
      a: {type: str, default: 0.3}
      wncutoff: {type: str, default: 0.000694}
      ycutoff: {type: str, default: 3}
      ydcutoff: {type: str, default: 30}
      shap_data_size: {type: str, default: 10}
      analyze_with_shap: {type: str, default: false}
      multiple: {type: str, default: "false"}
      eval_series: {type: str, default: "Portugal"}
      n_trials: {type: str, default: 100}
      opt_test: {type: str, default: "false"}
      from_mongo: {type: str, default: "false"}
      mongo_name: {type: str, default: "rdn_load_data"}
      num_workers: {type: str, default: 4}
      eval_method: {type: str, default: "ts_ID"}
      l_interpolation: {type: str, default : "false"}
      rmv_outliers: {type: str, default: "true"}
      loss_function: {type: str, default: "mape"}
      evaluate_all_ts: {type: str, default: "false"}
      convert_to_local_tz: {type: str, default: "true"}
      grid_search: {type: str, default: "false"}
      input_chunk_length: {type: str, default: None}
      ts_used_id: {type: str, default: "None"}
      m_mase: {type: str, default: "1"}
      min_non_nan_interval: {type: str, default: "24"}
      num_samples: {type: str, default: "1"}

    command: |
      python ../experimentation_pipeline.py --series-csv {series_csv} --series-uri {series_uri} --resolution {resolution} --year-range {year_range} --time-covs {time_covs} --cut-date-val {cut_date_val} --cut-date-test {cut_date_test} --test-end-date {test_end_date} --darts-model {darts_model} --device {device} --hyperparams-entrypoint {hyperparams_entrypoint} --forecast-horizon {forecast_horizon} --stride {stride} --retrain {retrain} --ignore-previous-runs {ignore_previous_runs} --scale {scale} --scale-covs {scale_covs} --day-first {day_first} --country {country} --std-dev {std_dev} --max-thr {max_thr} --a {a} --wncutoff {wncutoff} --ycutoff {ycutoff} --ydcutoff {ydcutoff} --shap-data-size {shap_data_size} --analyze-with-shap {analyze_with_shap} --multiple {multiple} --eval-series {eval_series} --n-trials {n_trials} --opt-test {opt_test} --from-mongo {from_mongo} --mongo-name {mongo_name} --num-workers {num_workers} --eval-method {eval_method} --l-interpolation {l_interpolation} --rmv-outliers {rmv_outliers} --loss-function {loss_function} --evaluate-all-ts {evaluate_all_ts} --convert-to-local-tz {convert_to_local_tz} --grid-search {grid_search} --input-chunk-length {input_chunk_length} --ts-used-id {ts_used_id} --m-mase {m_mase} --min-non-nan-interval {min_non_nan_interval} --past-covs-csv {past_covs_csv} --past-covs-uri {past_covs_uri} --future-covs-csv {future_covs_csv} --future-covs-uri {future_covs_uri} --num-samples {num_samples}

  inference:
    parameters:
      pyfunc_model_folder: {type: str, default: s3://mlflow-bucket/2/33d85746285c42a7b3ef403eb2f5c95f/artifacts/pyfunc_model}
      forecast_horizon:  {type: str, default: 960}
      series_uri: {type: str, default: series.csv}
      past_covariates_uri: {type: str, default: "None"}
      future_covariates_uri: {type: str, default: "None"}
      roll_size: {type: str, default: 96}
      batch_size:  {type: str, default: 1}

    command: |
      python ../inference.py --pyfunc-model-folder {pyfunc_model_folder} --forecast-horizon {forecast_horizon} --series-uri {series_uri} --past-covariates-uri {past_covariates_uri} --future-covariates-uri {future_covariates_uri} --roll-size {roll_size} --batch-size {batch_size}

\end{minted}

\section{Forecasting back-end architecture} \label{app:b}

The architecture of the forecasting back-end is illustrated in Fig. \ref{fig:mlflow_architecture}:

\begin{figure}[H]
    \centering
    \includegraphics[width=\textwidth]{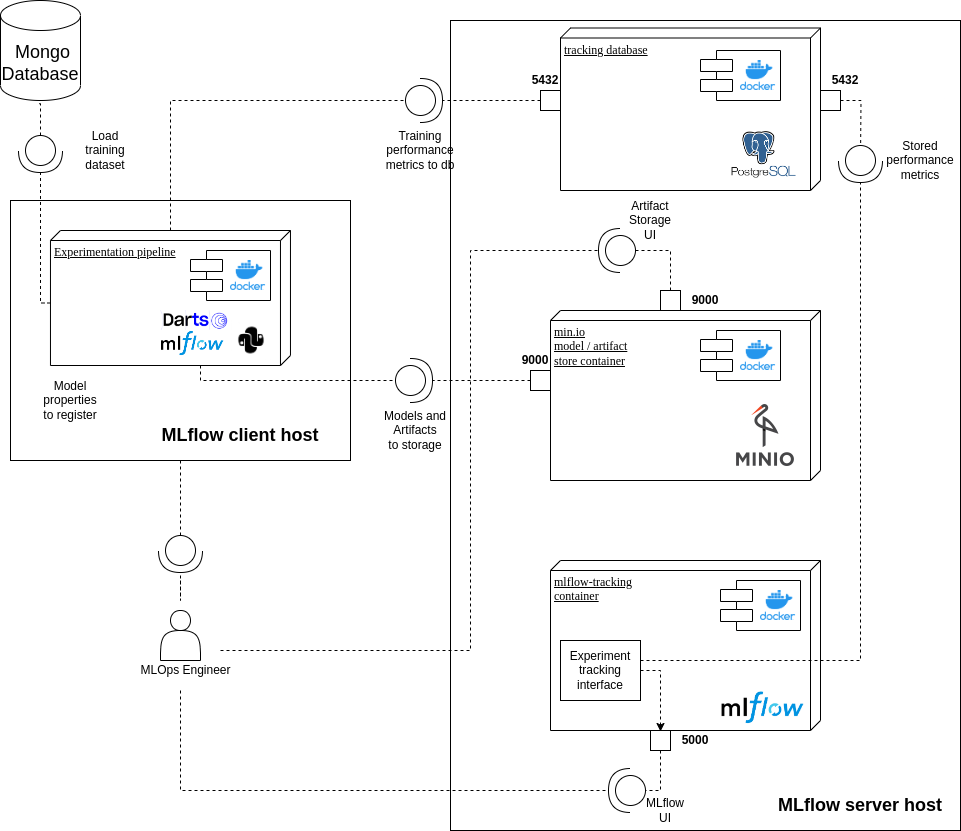}
    \caption{MLflow architecture diagram}
    \label{fig:mlflow_architecture}
\end{figure}

The components that run on the MLflow server host are responsible for the storage of the models and of other relevant artifacts (MinIO model / artifact store container), for the logging of parameters and metrics of each step (tracking database), as well as for running the MLflow user interface presented to the user (MLflow tracking container). The component on the left (which runs on the MLflow client host) is the executor the DeepTSF ML pipelines which acts as a client to the services of the MLflow server. All these components are containerized using Docker and can either run on the same ore different machines.

\section{Permitted file format}
\label{app:c}
The format of the csv files DeepTSF can accept depends on the nature of the problem it is trying to solve. More specifically, in case of a single time series file, its format is illustrated in Table \ref{tab:csv_simple}:

\begin{table}[H]
\centering
\caption{Single time series file format in hourly resolution}
\begin{tabular}{l|l}
\hline
Datetime                                            & Value  \\
\hline
2015-04-09 00:00:00                             & 7893 \\
2015-04-09 01:00:00               & 8023 \\
2015-04-09 02:00:00   & 8572 \\
...                                             & ... \\
\hline
\end{tabular}
\label{tab:csv_simple}
\end{table}

In this table, the Datetime column simply stores the dates and times of each observation, and the Value column stores the value that has been observed.
\par 

If we are solving a multiple and / or multivariate time series problem, then the file format (along with example values) is shown in Table \ref{tab:csv_multi}:

\begin{table}[hbt!]
\centering
\caption{Multiple and / or multivariate time series file format}
\begin{tabular}{l|l|l|l|l|l|l|l}
\hline
Index & Date & ID & Timeseries ID & 00:00:00 & ... \\ 
\hline
0     & 2015-04-09  & PT  & PT & 5248 & ... \\
1     & 2015-04-09  & ES  & ES & 25497 & ...\\
...   & ...  & ...  & ... & ... & ... \\
\hline
\end{tabular}
\label{tab:csv_multi}
\end{table}
\par 

The columns that can be present in the csv have the following meaning:
\begin{itemize}
    \item Index: Simply a monotonic integer range
    \item Date: The Date each row is referring to
    \item ID: Each ID corresponds to a component of a time series in the file. This ID must be unique for each time series component in the file. If referring to country loads it can be the country code. In this case, this will be used to obtain the country holidays for the imputation function as well as the time covariates.
    \item Timeseries ID (Optional): Timeseries ID column is not compulsory, and shows the time series to which each component belongs. If Timeseries ID is not present, it is assumed that each component represents one separate series (the column is set to ID).
    \item Time columns: Columns that store the Value of each component. They must be consecutive and separated by resolution minutes. They should start at 00:00:00, and end at 24:00:00 - resolution
\end{itemize}
\par

The checks that are performed when valifating a file are the following:

For all time series:
\begin{itemize}
    \item The dataframe can not be empty
    \item All the dates must be sorted
\end{itemize}

For non-multiple time series:
\begin{itemize}
    \item Column Datetime must be used as an index
    \item If the time series is the main dataset, Load must be the only other column in the dataframe
    \item If the time series is a covariates time series, there must be only one column in the dataframe named arbitrarily
\end{itemize}

For multiple timeseries:
\begin{itemize}
    \item Columns Date, ID, and the time columns exist in any order
    \item Only the permitted column names exist in the dataframe (see Multiple timeseries file format bellow)
    \item All timeseries in the dataframe have the same number of components
\end{itemize}

For more information about these files see the documentation \citep{Shap2023SHAPDocumentation}.

\section{Providing the hyperparameters to DeepTSF}
\label{app:d}

\par To use DeepTSF's optimization mechanism the user needs to provide the desired hyperparameter grid in the config\_opt.yml file using the YAML format \citep{YAML2023YAML1.2.2}, as shown in Fig. \ref{fig:hyperparams}. This is the grid used for the example of \ref{sec:3:CLI}. In the YAML file, the possible values for each hyperparameter need to be given in a list format as follows:
\begin{itemize}
\item Format ["range", start, end, step]: a list of hyperparameter values are considered ranging from value "start" till "end" with the step being defined by the last value of the list. 
\item Format ["list", value\_1, ..., value\_n]: All the listed parameters (\{value\_1, ..., value\_n\}) are considered in the grid. 
\end{itemize}
More information is included in the documentation of DeepTSF \citep{DeepTSF2023DeepTSFDocumentation}.

\begin{figure}[H]
\begin{minted}[fontsize=\footnotesize, linenos=False, frame=lines, framesep=2mm]{yaml}
NBEATS_example:
    input_chunk_length: ["range", 48, 240, 24]
    output_chunk_length: 24
    num_stacks: ["range", 1, 10, 1]
    num_blocks: ["range", 1, 10, 1]
    num_layers: ["range", 1, 5, 1]
    generic_architecture: True
    layer_widths: 64
    expansion_coefficient_dim: 5
    n_epochs: 300
    random_state: 0
    nr_epochs_val_period: 2
    batch_size: ["list", 256, 512, 1024, 1280, 1536, 2048]
\end{minted}
\caption{Hyperparameter tuning values of N-BEATS model}
\label{fig:hyperparams}
\end{figure}

\end{document}